\documentclass[letterpaper, 10 pt, conference]{ieeeconf}  

\IEEEoverridecommandlockouts                              

\usepackage{graphics} 
\usepackage{epsfig} 
\usepackage{times} 
\usepackage{amsmath} 
\usepackage{amssymb}  
\usepackage{hyperref}       
\usepackage{color}
\usepackage{adjustbox}
\usepackage{stfloats}
\usepackage{subcaption}
\usepackage{algorithm}
\usepackage{algorithmic}

\usepackage{titlesec}
\titleformat{\paragraph}
  [runin] 
  {\normalfont\bfseries} 
  {} 
  {-0.5em} 
  {} 

\title{\LARGE \bf
{
VividFace: Real-Time and Realistic Facial Expression Shadowing \\ for Humanoid Robots}
}

\author{Peizhen Li$^{1}$, Longbing Cao$^{1}$, Xiao-Ming Wu$^{2}$, and Yang Zhang$^{3}$
\thanks{$^{1}$Peizhen Li and Longbing Cao are with the School of Computing, Macquarie University, Australia
{\tt\small (peizhen.li1@hdr.mq.edu.au; longbing.cao@mq.edu.au)}.}%
\thanks{$^{2}$Xiao-Ming Wu is with the College of Computing and Data Science, Nanyang Technological University, Singapore
        {\tt\small (xiaoming.wu@ntu.edu.sg)}.}%
\thanks{$^{3}$Yang Zhang is with the Anuradha and Vikas Sinha Department of Data Science, University of North Texas, USA
        {\tt\small (yang.zhang@unt.edu)}.}%
}

\begin{document}

\maketitle
\thispagestyle{empty}
\pagestyle{empty}


\begin{abstract}
Humanoid facial expression shadowing enables robots to realistically imitate human facial expressions in real time, which is critical for lifelike, facially expressive humanoid robots and affective human–robot interaction. Existing progress in humanoid facial expression imitation remains limited, often failing to achieve either real-time performance or realistic expressiveness due to offline video-based inference designs and insufficient ability to capture and transfer subtle expression details.
To address these limitations, we present VividFace, a real-time and realistic facial expression shadowing system for humanoid robots.
An optimized imitation framework X2CNet++ enhances expressiveness by fine-tuning the human-to-humanoid facial motion transfer module and introducing a feature-adaptation training strategy for better alignment across different image sources. 
Real-time shadowing is further enabled by a video-stream-compatible inference pipeline and a streamlined workflow based on asynchronous I/O for efficient communication across devices.
VividFace produces vivid humanoid faces by mimicking human facial expressions within 0.05 seconds, while generalizing across diverse facial configurations. Extensive real-world demonstrations validate its practical utility. Videos are available at: https://lipzh5.github.io/VividFace/.

\end{abstract}


\section{Introduction}
Facial expressions are a primary channel for conveying emotions and social signals in human communication~\cite{mehrabian2017communication}. For humanoid robots, the ability to imitate human facial expressions realistically and in real time is crucial for achieving affective and engaging human–robot interaction (HRI)~\cite{li2024ugotme, cao2024humanoid}. By mirroring human expressions, robots can foster empathy~\cite{davis2018empathy}, build trust, and enhance collaboration in applications ranging from social assistance to education and healthcare.

Recent advances have enabled notable progress in humanoid facial expression imitation~\cite{li2025x2c,chen2021smile,li2023design,hu2024human,liu2024unlocking}, yet existing methods typically fall short of achieving both real-time performance and realistic expressiveness. Some approaches prioritize speed but oversimplify expression representations, making them unable to capture subtle details such as frowns and wrinkles in humans and transfer them to humanoids, while others capture fine-grained nuances but are computationally intensive or incompatible with real-time interaction. Bridging these two requirements remains an open challenge.

\begin{figure}[h]
    \centering
    \includegraphics[width=1.0\linewidth]{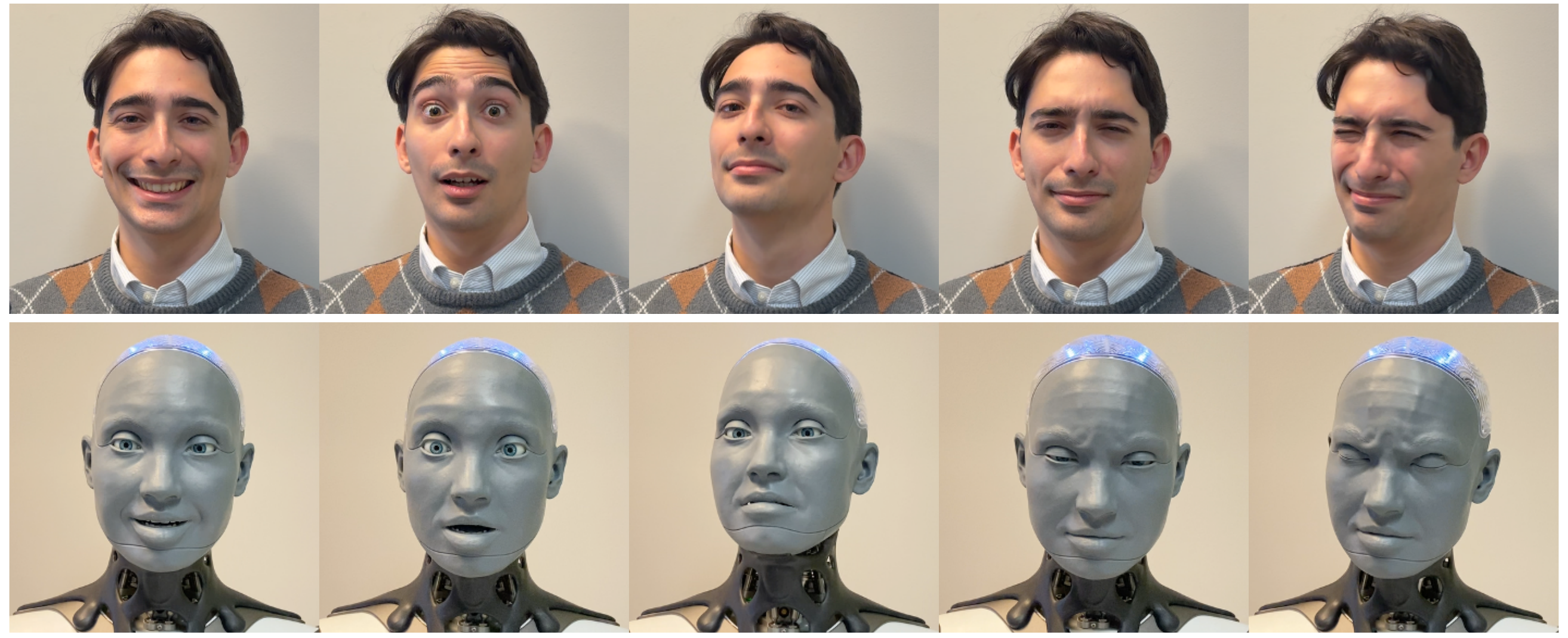}
    \caption{A demonstration of \textbf{VividFace}. The humanoid robot faithfully imitates the facial expressions of the human performer in {real time}. The \textbf{shadowing of subtle details}, such as frowning, gaze direction, and head pose enhances realism.}
    \label{fig:demonstration}
\end{figure}

\begin{figure*}
    \centering
    \includegraphics[width=1.\linewidth]{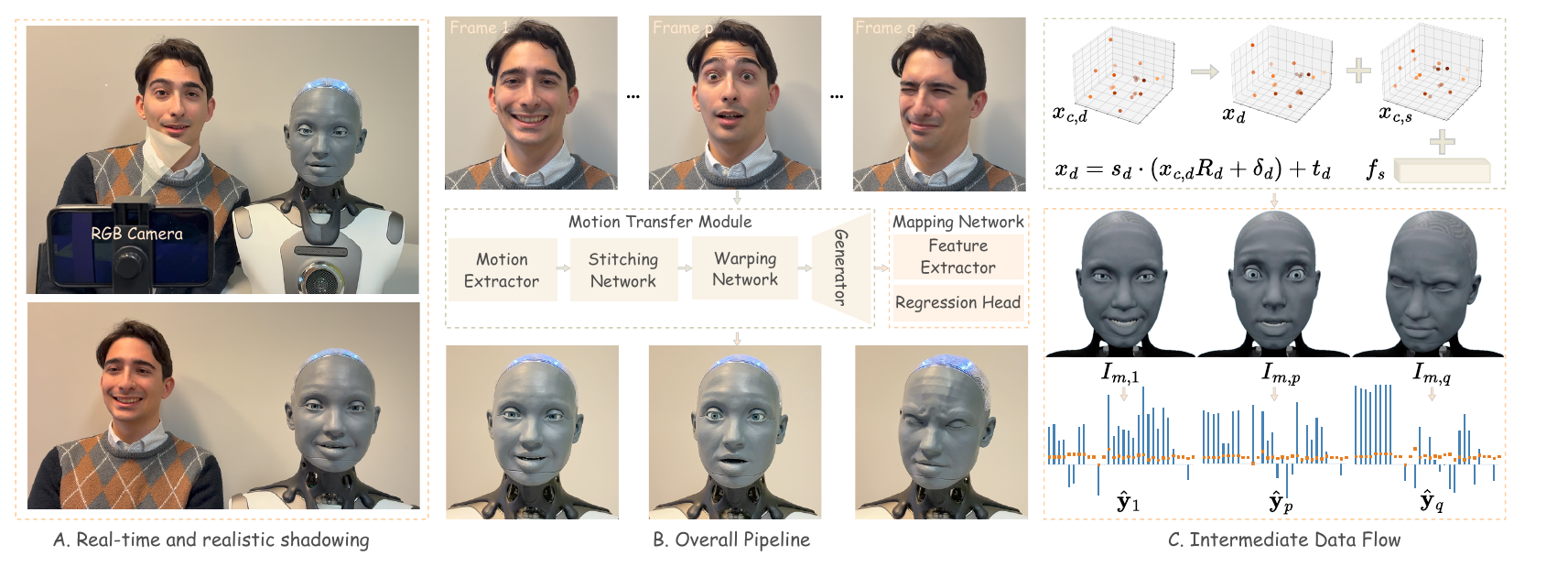}
    \caption{An overview of the \textbf{VividFace} workflow. An RGB camera captures human facial expression dynamics (A), and the image frames (each frame denoted by $I_d$) are streamed to the server and processed by the imitation framework, which consists of the motion transfer module $\mathcal{M}_1$ and the mapping network $\mathcal{M}_2$. The motion transfer module produces an intermediate expression representation $I_m = \mathcal{M}_1(I_d; f_s, x_{c,s})$ that integrates human motion with a virtual robot face. The mapping network then predicts control values $\hat{\mathbf{y}} = \mathcal{M}_2(I_m)$, which are used to drive the physical robot to reproduce the expression (B). The intermediate data flow for three example frames is visualized on the right (C).
    }
    \label{fig:sysoverview}
\end{figure*}

To address this open challenge, we propose \textbf{VividFace}, a real-time and realistic facial expression shadowing system for humanoid robots. As illustrated in Fig.~\ref{fig:demonstration}, \textbf{VividFace} enables robots to faithfully imitate the facial expressions of human performers in real time. Subtle details such as frowning, gaze direction, and head pose are shadowed, ensuring the synchronization of emotional nuances and enhancing the realism of humanoid expressions. {A video demonstration corresponding} to Fig.~\ref{fig:demonstration} is available at: https://lipzh5.github.io/VividFace/.
 An overview of the \textbf{VividFace} workflow is shown in Fig.~\ref{fig:sysoverview}. An RGB camera captures the dynamics of human facial expressions (A) and streams image frames (each denoted by $I_d$) to the server, where they are processed by the two-stage imitation framework.
In the first stage, the \textbf{human-to-humanoid facial motion transfer} (or simply \textbf{motion transfer}) module $\mathcal{M}_1$ combines the implicit-keypoint-based human facial motion with the humanoid facial features $f_{s}$ and keypoints $x_{c,s}$ to generate an intermediate expression representation $I_m $ in image space. In the second stage, the mapping network $\mathcal{M}_2$ projects $I_m$ to control values $\hat{\textbf{y}} = \mathcal{M}_2(I_m)$, which drive the physical robot to reproduce the expressions (B).
The intermediate data flow for frames 1, $p$, and $q$ is visualized in C, where $x_{c,d}$ and $x_{c,s}$ denote the facial keypoints of a driving frame (displaying a human expression) and a source frame (displaying a humanoid expression), respectively. Further details on the facial keypoint transformation are provided in Sec.~\ref{sec:preliminary}.

We observe that the motion transfer module, which is primarily pretrained on human facial images, sometimes fails to faithfully transfer expression nuances—such as wrinkles—from the human face to the humanoid face (Fig.~\ref{fig:wrinklemissing}). To address this limitation, we fine-tune the module on the X2C dataset~\cite{li2025x2c}, a large-scale dataset featuring nuanced humanoid facial expressions, by formulating a self-supervised image reconstruction task and adopting the Generative Adversarial Network (GAN) training paradigm~\cite{guo2024liveportrait,siarohin2019first}.

However, the generated intermediate expression representation $I_m$—the input to the mapping network $\mathcal{M}_2$ during real-time shadowing—may still differ from the humanoid facial images used to train $\mathcal{M}_2$ (see input images $I_x$ and $\tilde{I}_x$ in Fig.~\ref{fig:featurematching}). To bridge this gap, we introduce a feature-adaptation training strategy~\cite{ganin2016domain}, which exposes the model to inputs from multiple sources (the original training set and generated images simulating inference inputs) and aligns their features.
This encourages images conveying the same expression nuances to be aligned in feature space, regardless of domain or pixel-level differences. An illustration of this training process is provided in Fig.~\ref{fig:featurematching}, with further details in Sec.~\ref{sec: domain_adaptation}.

To enable real-time shadowing, we optimize human facial motion extraction for compatibility with video streaming and implement a streamlined workflow based on asynchronous I/O. This allows the robot to mimic human facial expressions within 0.05 seconds, achieving vivid and realistic shadowing~\cite{forch2017100}.
We conduct extensive real-world experiments on a physical humanoid robot with multiple human participants exhibiting diverse facial configurations, validating the practical utility and potential of \textbf{VividFace} for enhancing natural human–robot interaction.
Our main contributions include:
\begin{itemize}
\item \textbf{VividFace}: A real-time and realistic facial expression shadowing system enabling lifelike facially expressive humanoid robots.
\item \textbf{X2CNet++}: An optimized imitation framework that fine-tunes the motion transfer module for enhanced expression fidelity, with a feature-adaptation training strategy to bridge the gap between training and inference domains.
\item 
A streamlined imitation pipeline compatible with real-time streamed video data, achieving expression mimicry within 0.05 seconds.
\item Extensive real-world demonstrations, with code and model checkpoints released publicly, validating the effectiveness and generalizability of our approach.
\end{itemize}


\section{Related work}
\textbf{Humanoid Facial Expression Imitation.} Methods for humanoid facial expression imitation can be broadly divided into rule-based and learning-based approaches. Traditional methods have mainly relied on rule-based or preprogrammed expressions~\cite{kundu2016roboticon,cid2013real, esfandbod2019human, ge2008facial, kim2006development}. While simple and interpretable, these approaches restrict variability and result in constrained expressiveness. More recently, learning-based methods have emerged that imitate human expressions by training models on facial data, going beyond predefined categories~\cite{li2025x2c, liu2024unlocking}. For example X2CNet~\cite{li2025x2c} learns to capture fine-grained expression details from the human face and predict the control values for a humanoid face, leading to realistic humanoid imitation without predefined emotion category constraints. However, despite advances in realism and nuance, most approaches struggle to achieve real-time performance. We address this issue by introducing \textbf{VividFace}, a real-time expression shadowing system that integrates a streamlined imitation pipeline.

\textbf{Real-Time Facial Expression Imitation for Affective HRI.} 
Facial expressions are among the most effective modalities for communicating affective information compared to verbal cues and tone~\cite{mehrabian2017communication,li2024ugotme,fukuda2002facial,hyung2018facial}. 
Real-time imitation of human facial expressions is crucial for affective human–robot interaction~\cite{meghdari2016real, chen2021smile, hu2024human,liu2022real}. A facial landmark-based learning framework is proposed in~\cite{chen2021smile} that enables a humanoid robot to provide a mimicry response within 0.18 seconds. However, it fails to imitate subtle expression details such as nose wrinkles and frowning, as well as the accurate intensity of mouth and eye openness. \textbf{VividFace} addresses this issue by proposing an optimized imitation framework \textbf{X2CNet++}.
By achieving both high responsiveness ($<0.05$ seconds) and nuanced expressiveness, \textbf{VividFace} enhances natural, affective HRI and fosters more human-like engagement with robots.

\section{Method}
\begin{algorithm}[t]
\caption{Real-Time \& Realistic Expression Shadowing}
\label{alg:shadowing}
\begin{algorithmic}[1]
    \REQUIRE {Human face video stream} $\{I_{d}\}$, source humanoid face $I_s$, imitation framework (motion transfer module $\mathcal{M}_1$, mapping network $\mathcal{M}_2$)
    \ENSURE Real-time control values $\{\hat{\mathbf{y}}\}$ for humanoid robot
    
    \STATE Precompute source keypoints $x_{c,s}$ and feature volumes $f_s$ from $I_s$
    
    \FOR{each incoming driving frame $I_{d}$}
        \STATE Generate motion-transferred humanoid image: $I_m = \mathcal{M}_1(I_{d}; f_s, x_{c,s})$
        \STATE Predict control values: $\hat{\mathbf{y}} = \mathcal{M}_2(I_m)$
        \STATE Send $\hat{\mathbf{y}}$ to the humanoid robot
    \ENDFOR
\end{algorithmic}
\end{algorithm}

The primary contribution of this work is \textbf{VividFace}, a real-time and realistic facial expression shadowing system that enables humanoid robots to faithfully imitate human expressions, capturing subtle details such as wrinkles, frowning, and gaze direction. The overall workflow of \textbf{VividFace} is outlined in Algorithm~\ref{alg:shadowing}. To achieve real-time response, we streamline the imitation pipeline to ensure compatibility with streamed video data (Sec.~\ref{sec:realtime_processing}). To preserve expression fidelity, we incorporate the following optimizations to the imitation framework:
we fine-tune the motion transfer module on the X2C dataset by formulating an image reconstruction task following the GAN training paradigm (Sec.~\ref{sec:finetune}) and introduce a feature-adaptation training strategy for the mapping network to bridge the gap between training and inference domains (Sec.~\ref{sec: domain_adaptation}). The optimized imitation framework is denoted as \textbf{X2CNet++}.

\subsection{Preliminaries on Motion Transfer and Humanoid Facial Control}
\label{sec:preliminary}
\textbf{Implicit-keypoint-based facial motion transfer.} To achieve realistic imitation without sacrificing computational efficiency, we employ the implicit-keypoint-based method LivePortrait~\cite{guo2024liveportrait} for the motion transfer module, following~\cite{li2025x2c}. Here, we briefly introduce the basic idea of the implicit-keypoint-based motion transfer algorithm.

Given a driving image $I_d$ that provides facial motion information and a source image $I_s$ that provides appearance information, the motion extractor estimates canonical 3D keypoints $x_{c,s}, x_{c,d} \in \mathbb{R}^{K \times 3}$, head poses $R_s, R_d \in \mathbb{R}^{3 \times 3}$, expression deformations $\delta_s, \delta_d \in \mathbb{R}^{K \times 3}$, translations $t_s, t_d \in \mathbb{R}^{3}$, and scale factors $s_s, s_d$ for both source and driving images. Here, $K$ denotes the number of keypoints. The keypoint transformations are formulated as: 

\begin{equation}
\begin{cases}
x_s = s_s \cdot (x_{c,s} R_s + \delta_s) + t_s, \\
x_d = s_d \cdot (x_{c,d} R_d + \delta_d) + t_d.    
\end{cases}
\end{equation}
The stitching network $\mathcal{S}$ estimates the deformation offset 
\[
\Delta = \mathcal{S}(x_s, x_d) \in \mathbb{R}^{K \times 3},
\]
which is used to update the driving keypoints: 
\[
x^{\prime}_d = x_d + \Delta.
\]
The warping network $\mathcal{W}$ generates a warping field using $x_s$ and $x^{\prime}_d$, which is then applied to the source feature volume $f_s$. The warped feature passes through a generator $\mathcal{G}$, producing an image with the driving motion and source appearance:
\[
I_m = \mathcal{G}(\mathcal{W}(f_s; x_s, x^{\prime}_d)).
\]
For \textbf{VividFace}, the motion extractor, stitching network, warping network, appearance feature extractor and the generator are all encapsulated into the motion transfer module $\mathcal{M}_1$. Thus, we can express it as
\[
I_m = \mathcal{M}_1(I_d; f_s, x_{c,s}).
\]

\textbf{Control values and the humanoid facial expressions.} Each humanoid facial expression corresponds to a sequence of control values $\mathbf{y} \in \mathbb{R}^C$. 
We employ a mapping network $\mathcal{M}_2$ to learn the correspondence between intermediate expression representations and {control values}, following X2CNet~\cite{li2025x2c}:
\[
    \hat{\mathbf{y}} = \mathcal{M}_2(I_m),
\]
where $\hat{\mathbf{y}} \in \mathbb{R}^C$ are the predicted control values.
Visualizations of control values ($\hat{\mathbf{y}}_1$, $\hat{\mathbf{y}}_p$, and $\hat{\mathbf{y}}_q$) for three example expressions are provided in Fig.~\ref{fig:sysoverview} (C. Intermediate Data Flow), 
where blue bars indicate the control values and orange dots denote the neutral-state reference. 
Applying these control values to the physical robot reproduces the corresponding intermediate expressions $I_{m,1}$, $I_{m,p}$, and $I_{m,q}$, 
as illustrated at the bottom of Fig.~\ref{fig:sysoverview} (B. Overall Pipeline).

\subsection{Video Streaming and Real-Time Processing}
\label{sec:realtime_processing}
The input to the system is a video stream capturing human facial expression dynamics using the RGB camera of an iPhone 11. We developed a customized iOS application in Xcode to stream the captured frames to a local server ({the application source code} is available at: https://github.com/lipzh5/VividFace).
Data transmission is performed via the HTTP protocol, with image compression quality set to 0.8 to balance speed and image quality. This configuration yields an average frame rate of around 30 FPS and an image resolution of $480 \times 360$ at the server side.

This streaming design is low-cost and flexible, allowing adjustments of both frame rate and image resolution. 
During real-time expression shadowing, given a human face video stream $\{I_{d,i} \mid i=0, \ldots, N-1\}$ and a humanoid source image $I_s$, 
we condition the driving facial keypoints on the source humanoid keypoints and transform them using relative motion. 
Specifically, we define the relative transformation of keypoints $x_{d,i}$ for each driving frame as follows:
\begin{equation}
     x_{d,i} = \bar{s}_{d,i} \cdot 
      (x_{c,s} \bar{R}_{d,i} 
      + \bar{\delta}_{d,i}) + \bar{t}_{d,i},
\end{equation}
where 
\[\bar{s}_{d,i} = s_s \tfrac{s_{d,i}}{s_{d,0}}, ~ \bar{R}_{d,i} = R_{d,i} R_{d,0}^{-1} R_s, ~\bar{\delta}_{d,i} = \delta_s + \delta_{d,i} - \delta_{d,0},\]
and $\bar{t}_{d,i} = t_s + t_{d,i} - t_{d,0}$ 
denote the relative scale factor, rotation, expression deformation, and translation with respect to the source humanoid frame and the first driving human frame, respectively.
To improve efficiency, we apply the following optimizations:
\begin{enumerate}
    \item Cache the source canonical keypoints and feature volumes instead of recalculating them for every driving frame, since they are always extracted from the same humanoid facial image during real-time expression shadowing.
   
     \item Implement a streaming version of the facial motion transfer. Existing offline video-based imitation methods assume access to all driving frames in advance, which is unsuitable for real-time applications. Instead, we compute the keypoint transformation and feature warping for each incoming frame, enabling immediate generation of the intermediate expression representation and prediction of control values. This design improves the smoothness and responsiveness of expression shadowing.

    \item Employ asynchronous I/O for communication between the server and the robot, thereby avoiding unnecessary blocking, improving resource utilization, and reducing stutter during on-robot execution. This optimization enables the robot to deliver stable, low-latency mimicry responses even under fluctuating input frame rates.

\end{enumerate}
With these optimizations, the humanoid robot can mimic human facial expressions within 0.05 seconds, achieving a natural real-time shadowing effect.



\subsection{Fine-tuning the Motion Transfer Module}
\label{sec:finetune}


\begin{figure}[htbp]
  \centering
  
  \begin{minipage}{0.99\linewidth}
  \begin{subfigure}{1.0\linewidth}
    \centering
    \includegraphics[width=\linewidth]{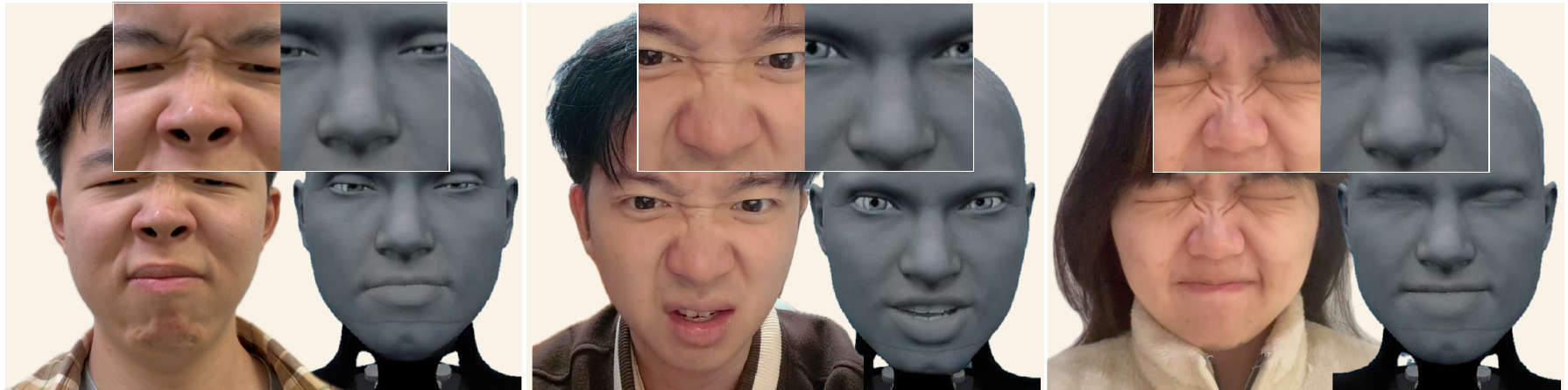}
    \caption{Failure cases where nose wrinkles are \textbf{not transferred} from the human face to the robot face.}
    \label{fig:wrinklemissing}
  \end{subfigure}
  \end{minipage}
  
  \vspace{1em} 
  \begin{minipage}{0.99\linewidth}
  \begin{subfigure}{1.0\linewidth}
    \centering
    \includegraphics[width=\linewidth]{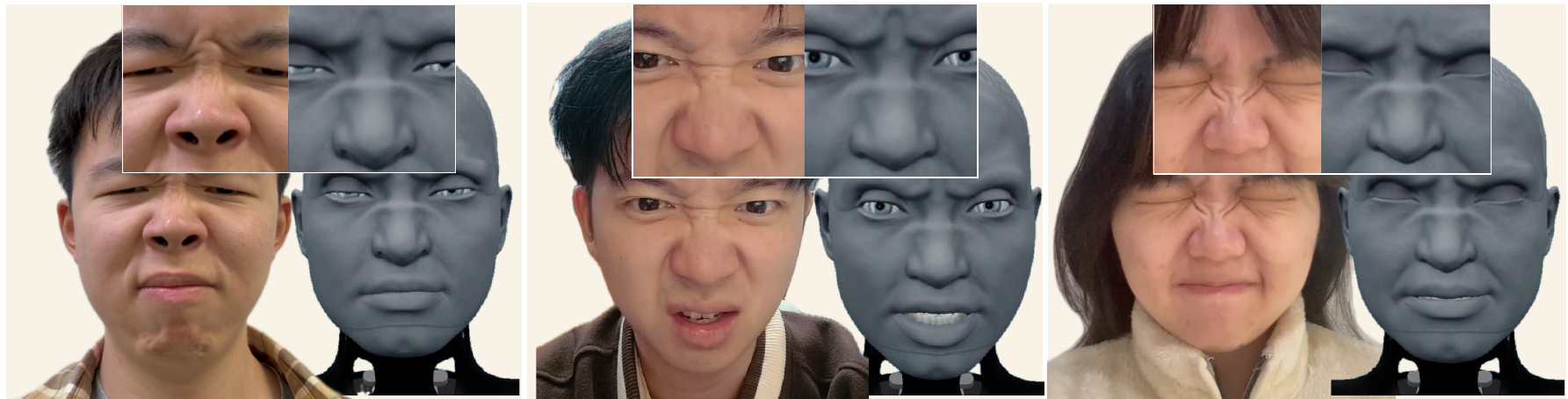}
    \caption{With motion transfer module fine-tuning, nose wrinkles are \textbf{successfully transferred} to the robot face.}
    \label{fig:wrinkles_solved}
  \end{subfigure}
  \end{minipage}
  \caption{Comparison of nose wrinkle transfer in models with and without fine-tuning on the X2C dataset.}
  \label{fig:comparison_wrinkles}
\end{figure}

We notice that some subtle expression details, such as nose wrinkles, cannot be transferred from human faces to the virtual robot’s face (Fig.~\ref{fig:wrinklemissing}), which serves as our intermediate expression representation used to predict control values for driving the physical robot. Therefore, the absence of wrinkles in the intermediate expression representation propagates to the physical robot.

The key reason is that the original motion transfer module is pretrained on datasets consisting of human faces, such as Voxceleb~\cite{nagraniy2017voxceleb} and MEAD~\cite{wang2020mead}, which works well in transferring expression dynamics between human faces but may not generalize well to unseen humanoid faces, since they differ significantly at the pixel level. 
To address this issue, we fine-tune the base models in the motion transfer module~\cite{guo2024liveportrait}—including the appearance feature extractor, motion extractor, warping network, and generator—using humanoid facial images from the X2C dataset~\cite{li2025x2c}. This is formulated as an image reconstruction task.

Following~\cite{guo2024liveportrait}, we fine-tune the models using the Hinge GAN loss, along with perceptual and feature-matching losses. We adopt a multiscale discriminator as in~\cite{siarohin2019first} and use features from a pretrained VGG-19 network for perceptual loss~\cite{johnson2016perceptual}. Additionally, we incorporate a feature-matching loss to stabilize adversarial training and encourage the generator to capture fine details, as suggested in~\cite{mescheder2018training}.
The GAN training objectives for image reconstruction are formulated as follows:

\begin{equation}
\begin{aligned}
L_D &= \mathbb{E}_{I_x \sim p_{\text{data}}} \Big[ \max(0, 1 - D(I_x)) \Big] \\
      &+ \mathbb{E}_{\tilde{I}_{{x}} \sim p_G} \Big[ \max(0, 1 + D(\tilde{I}_{{x}})) \Big], \\
L_G &= - \mathbb{E}_{\tilde{I}_{{x}} \sim p_G} \big[ D(\tilde{I}_{{x}}) \big] 
      + \lambda_{\text{p}} L_{\text{p}} 
      + \lambda_{\text{fm}} L_{\text{fm}},
\end{aligned}
\end{equation}
where $I_x$ is a real image sampled from the data distribution $p_{\text{data}}$, $\tilde{I}_x$ is the generated/reconstructed image, and $p_G$ represents the distribution of generated images; $L_{\text{p}}$ denotes the perceptual loss computed using features from a pretrained VGG-19 network, $L_{\text{fm}}$ denotes the feature-matching loss using intermediate discriminator features, and $\lambda_{\text{p}}$ and $\lambda_{\text{fm}}$ are the corresponding weighting factors.

\textbf{Implementation details.} We set the number of keypoints to $K=21$, and the loss weights to $\lambda_\text{p} = 10$ and $\lambda_\text{fm} = 10$. The scales for the multiscale discriminator are set to $[1, 0.5, 0.25, 0.125]$. For all base models, we use the Adam optimizer with a learning rate of $2.0 \times 10^{-5}$ and $\beta = (0.5, 0.999)$. We adopt MultiStepLR as the learning rate scheduler with $\gamma = 0.1$. The models are fine-tuned for 30 epochs, with epoch milestones at $[12, 18]$. 


\subsection{Feature Adaptation for Mapping Network Training}
\label{sec: domain_adaptation}
\begin{figure}[h]
    \centering
    \includegraphics[width=1.0\linewidth]{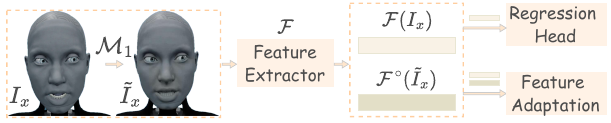}
    \caption{An illustration of the feature-adaptation training. $I_x$ refers to the image from the X2C dataset, while $\tilde{I}_x = \mathcal{M}_1(I_x)$ denotes its generated counterpart.}
    \label{fig:featurematching}
\end{figure}
The mapping network is originally trained on the X2C dataset to learn the correspondence between humanoid facial images and control values. However, during inference, it receives images \textbf{generated} by the motion transfer module—which may differ from the original training data (see example input images in Fig.~\ref{fig:featurematching})—as input. This gap between training and inference images can hinder the model from predicting accurate control values during real-world facial expression shadowing. To address this, we propose a \textbf{feature-adaptation training strategy} to better align features from different image sources~\cite{wei2021metaalign}, ensuring improved adaptability during inference and resulting in more authentic humanoid imitation.

An illustration of this training process is shown in Fig.~\ref{fig:featurematching}.
Specifically, for each image $I_x$ in X2C, we obtain its generated counterpart $\tilde{I}_x$ by passing it through the motion transfer module: $\tilde{I}_x = \mathcal{M}_1(I_x)$. 
During training, whenever we calculate the features for $I_x$: $f_{I_x} = \mathcal{F}(I_x)$, 
we immediately calculate the feature for $\tilde{I}_x$: $f_{\tilde{I}_x}^{\circ} = \mathcal{F}^{\circ}(\tilde{I}_x)$ where gradient tracking is disabled.
Thus, $f_{\tilde{I}_x}^{\circ}$ requires no gradient and serves as the target feature from another domain (similar to inference). We use the superscript $\circ$ to indicate features/tensors that do not require gradients during training. This ensures that both features are extracted by the same network with the same weights, while preventing the feature-adaptation step from interfering with control value prediction. The feature-adaptation loss is formulated as:
\begin{equation}
    L_{\text{fa}} = \|f_{I_x} - f^\circ_{\tilde{I}_x}\|_2^2,
\end{equation}
where $\|\cdot\|_2$ denotes the $L_2$ norm.
The regression loss is the Huber loss defined as:
\begin{equation}
    L_\delta(\mathbf{y}, \hat{\mathbf{y}}) = \frac{1}{C} \sum_{i=1}^{C} 
    \begin{cases} 
        \frac{1}{2} (y_i - \hat{y}_i)^2, & \text{if } |y_i - \hat{y}_i| \le \delta, \\
        \delta \, |y_i - \hat{y}_i| - \frac{1}{2} \delta^2, & \text{otherwise},
    \end{cases}
\end{equation}
where $\mathbf{y} = [y_1, \ldots, y_C]^\top \in \mathbb{R}^C$ denotes the ground-truth control vector, $\hat{\mathbf{y}} = [\hat{y}_1, \ldots, \hat{y}_C]^\top$ denotes the predicted control vector, and $\delta > 0$ is the threshold parameter.
The total training loss is then defined as: 
\begin{equation}
    L = L_\delta + \lambda_{\text{fa}} L_{\text{fa}},
\end{equation}
where $\lambda_{\text{fa}}$ is a weighting factor.

\textbf{Implementation details.} The total number of control values is $C = 30$. We set the threshold value to $\delta = 0.01$ and the weighting factor to $\lambda_{\text{fa}} = 5 \times 10^{-4}$. We use the AdamW optimizer with a learning rate of $0.001$ and apply a \textit{cosine schedule with warmup} as the learning rate scheduler. The batch size is set to 128, and the model is trained for 100 epochs. 








\section{Experiments}
In this section, we verify the effectiveness of the proposed method by comparing it with several baselines in terms of imitation realism and by reporting latency statistics for real-time evaluation. We also conduct ablation studies to assess the impact of fine-tuning the motion transfer module and the proposed feature-adaptation training strategy for the mapping network. Both qualitative and quantitative evaluations are included. For quantitative evaluation, we curate a test set of {2000} frames covering diverse human facial expressions across different facial configurations and recruit 5 human raters from varied cultural backgrounds. The humanoid robot used in the experiments is Ameca, which provides 32 Degrees of Freedom (DoF) for facial expression control. Model training and fine-tuning are performed on a single NVIDIA H100 GPU, while real-world deployment runs on a local server equipped with one NVIDIA RTX 4090 GPU.

\subsection{Baselines}
To validate the effectiveness of our proposed method in improving facial expression imitation realism, 
we compare the proposed \textbf{X2CNet++}, an optimized imitation framework that incorporates fine-tuning of the motion transfer module and feature-adaption training of the mapping network against the following baselines (for models without a specific name in the original papers, we select a word from the paper title as the model name):

\begin{enumerate}
\item \textbf{X2CNet}: a two-stage imitation framework introduced in~\cite{li2025x2c}, where the mapping network is trained on the X2C dataset using a pure regression loss. This model serves as the foundation for our proposed \textbf{X2CNet++}.
\item \textbf{Smile}: a vision-based self-supervised imitation framework consisting of a generative model and an inverse model~\cite{chen2021smile}, where the motor command values are discretized and the inverse model is trained using a multi-class cross-entropy loss.
\item \textbf{Coexpression}: a facial landmark-based inverse model for humanoid facial expression prediction proposed in~\cite{hu2024human}, which takes normalized facial landmarks as input and outputs a set of motor commands for controller execution.
\end{enumerate}

\subsection{Evaluation Metrics}

\textbf{Realism.} To evaluate facial expression imitation realism, we use both \textbf{objective} and \textbf{subjective} metrics:

\begin{itemize}
    \item \textbf{Mean Absolute Action Unit Intensity Difference (MAID):} an objective metric that measures the difference between intensities of Action Units (AUs) as defined in the Facial Action Coding System~\cite{ekman1978facial}. For the $t$-th expression pair (human vs. robot), the AU intensity vectors be $\mathbf{a}_t^h = [a_{t,1}^h, a_{t,2}^h, \dots, a_{t,n}^h]$ and $\mathbf{a}^r = [a_{t,1}^r, a_{t,2}^r, \dots, a_{t,n}^r]$, $a_{t,i}^h, a_{t,i}^r \in [0, 5]$, where $n$ is the number of AUs considered. MAID is calculated as:
    \begin{equation}
        \text{MAID} = \frac{1}{nT} \sum_{t=1}^T\sum_{i=1}^n \left| a_{t,i}^h - a_{t,i}^r \right|,
    \end{equation}
    where $T$ is the number of test samples. Lower values of MAID indicate higher similarity of facial expressions.
    \item \textbf{Average User Rating (AUR):} a subjective metric that captures human perception of imitation realism. A group of $m$ human raters are asked to watch paired expressions (human vs. robot) and rate the realism of the robot’s expression on a 5-point Likert scale (1 = very unrealistic, 5 = highly realistic). The AUR for a single rater is calculated as the mean score across all $T$ test samples:
    \begin{equation}
        \text{AUR} = \frac{1}{T} \sum_{t=1}^T r_{t},
    \end{equation}
where $r_{t}$ is the rating from a single rater for test sample $t$. Higher values of AUR indicate more realistic imitation. We report the mean and standard deviation across all $m$ raters.
\end{itemize}

\textbf{Real-time performance.} We use {end-to-end latency} to evaluate the real-time performance of the system, defined as the time interval between receiving the human motion input and sending facial expression-making command to the robot using predicted control values. 
Latency was measured under three CPU load conditions (idle, 50\%, 90\%), induced using \texttt{stress-ng}. 
Each condition was repeated three times (each lasting 10 minutes, approximately 10k samples).
CPU utilization was monitored with \texttt{mpstat} during the experiments. 
From the recorded latency samples, we compute standard statistics commonly used in real-time system evaluation~\cite{dean2013tail}:

\begin{itemize}
    \item \textbf{Mean latency:} the arithmetic average of all latency samples, representing typical system response.
    \item \textbf{P95 and P99 latency:} the 95th and 99th percentiles, representing tail latency. For example, the P95 latency is the value below which 95\% of the latency samples fall, capturing rare but important slow responses.
    \item \textbf{Maximum latency:} the largest observed latency among all samples, indicating the worst-case response.
    \item \textbf{Standard deviation (jitter):} the variability of latency around the mean.
\end{itemize}




\subsection{Results}
\subsubsection{Real-world Demonstrations}
\begin{figure*}[h]
    \centering
    \includegraphics[width=1.0\linewidth]{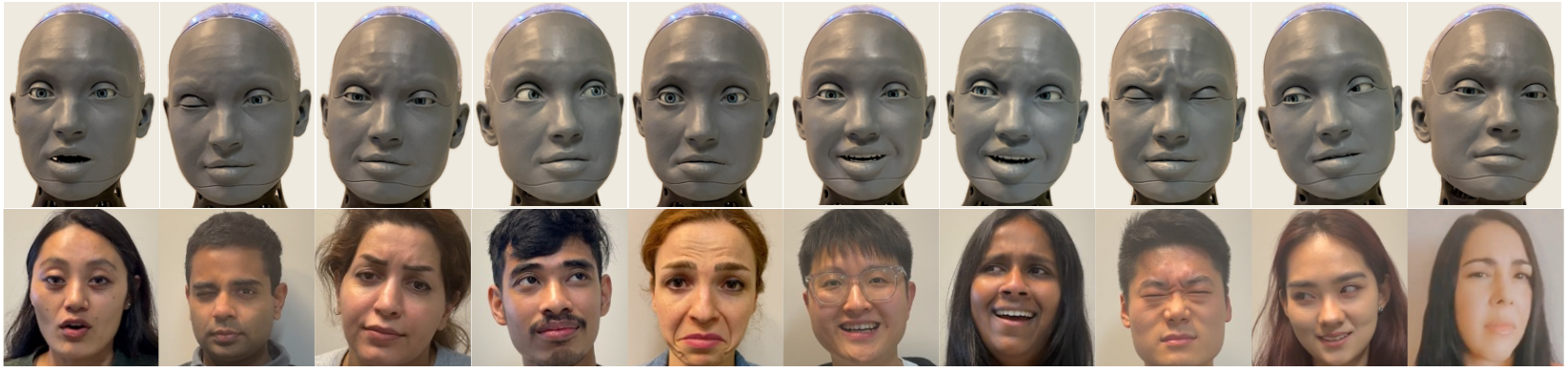}
    \caption{Real-world examples of humanoid robots performing realistic facial expression imitation.}
    \label{fig:realworld_examples}
\end{figure*}
We conduct extensive real-world experiments with human performers exhibiting diverse facial configurations to validate the generalizability and realism of our expression shadowing system. As shown in Fig.~\ref{fig:realworld_examples}, the robot can convincingly mimic a wide range of human facial expressions across variations in skin tone, hairstyle, and facial geometry. Notably, subtle details such as asymmetric winks (second from the left), frowns (third), gaze directions (fourth), nose wrinkles (eighth), and varying intensities of mouth and eye openness—which convey emotional nuance—are faithfully reproduced on the humanoid face. Our method also enables precise imitation of head pose, which is crucial for expressiveness (e.g., third and fourth). Additional video demonstrations showcasing real-time responsiveness and the shadowing effect are available on our project website: https://lipzh5.github.io/VividFace/.

\subsubsection{Benchmark Results}

\begin{figure}
    \centering
    \includegraphics[width=1.0\linewidth]{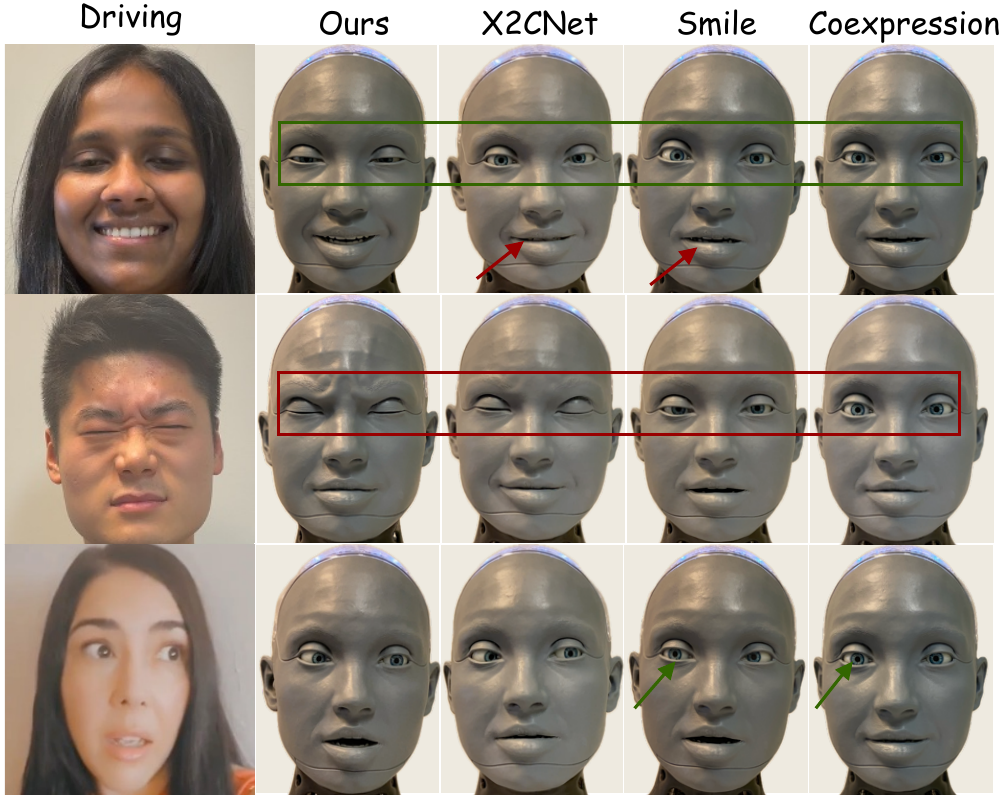}
    \caption{Qualitative comparisons of realistic humanoid imitation with baselines.}
    \label{fig:benchmark}
\end{figure}

\begin{table}[t]
\centering
\caption{Comparison with baseline methods in terms of AUR (1–5 scale) and {MAID}.}
\label{tab:benchmark}
\begin{tabular}{l c c}
\hline
\textbf{Method} & \textbf{AUR (Mean $\pm$ SD)}~$\uparrow$ &\textbf{MAID}~$\downarrow$\\
\hline
\textbf{X2CNet++} (Ours) & \textbf{4.76} $\pm$ {0.4027}  &\textbf{0.1810} \\
X2CNet~\cite{li2025x2c} & 3.53 $\pm$ 0.4988  & 0.2315 \\
Smile~\cite{chen2021smile}           & 2.23 $\pm$ 0.7498 &0.2698\\
Coexpression~\cite{hu2024human}             & 1.77 $\pm$ 0.7039  &0.2496\\
\hline
\end{tabular}
\end{table}

We compare the proposed \textbf{X2CNet++} with several baselines and present both qualitative and quantitative results in Fig.~\ref{fig:benchmark} and Table~\ref{tab:benchmark}. As shown in Fig.~\ref{fig:benchmark}, the proposed method outperforms the baselines by transferring more subtle details (e.g., nose wrinkles and frowning, as indicated by the red bounding box) and by producing more accurate openness of the eyes and mouth (green box and red arrows) as well as gaze directions (green arrows). For X2CNet, the shortcomings are twofold: (1) directly using the pretrained motion transfer module without fine-tuning on target humanoid facial data leads to the loss of expression details on the robot’s face; and (2) the mapping network (the second stage of the imitation framework), trained on the X2C dataset, may not generalize well to the input images during inference, which are the output images generated by the motion transfer module (the first stage of the imitation framework). For the other methods, where facial expression representations rely on facial landmarks, the limitations include: (1) the use of third-party tools such as OpenFace or OpenPose to detect landmarks introduces errors that propagate through subsequent stages and affect motor command prediction; and (2) fine-grained expression details are abstracted away by the landmark-based representations, resulting in inaccurate imitation of subtle openness and head pose variations.

For quantitative evaluation, we reported the AUR and MAID in Table~\ref{tab:benchmark} with Action Unit intensities extracted using OpenFace~\cite{baltruvsaitis2016openface}. As shown, the proposed \textbf{X2CNet++} enables the robot to imitate human facial expressions more realistically, achieving a mean AUR of 4.76, which indicates high imitation realism from human perception. For the objective metric MAID, our method achieves 0.1810, lower than all baselines, indicating greater similarity between human and robot expressions.

\subsubsection{Ablation Studies}
\begin{figure}[h]
    \centering
    \includegraphics[width=1.045\linewidth]{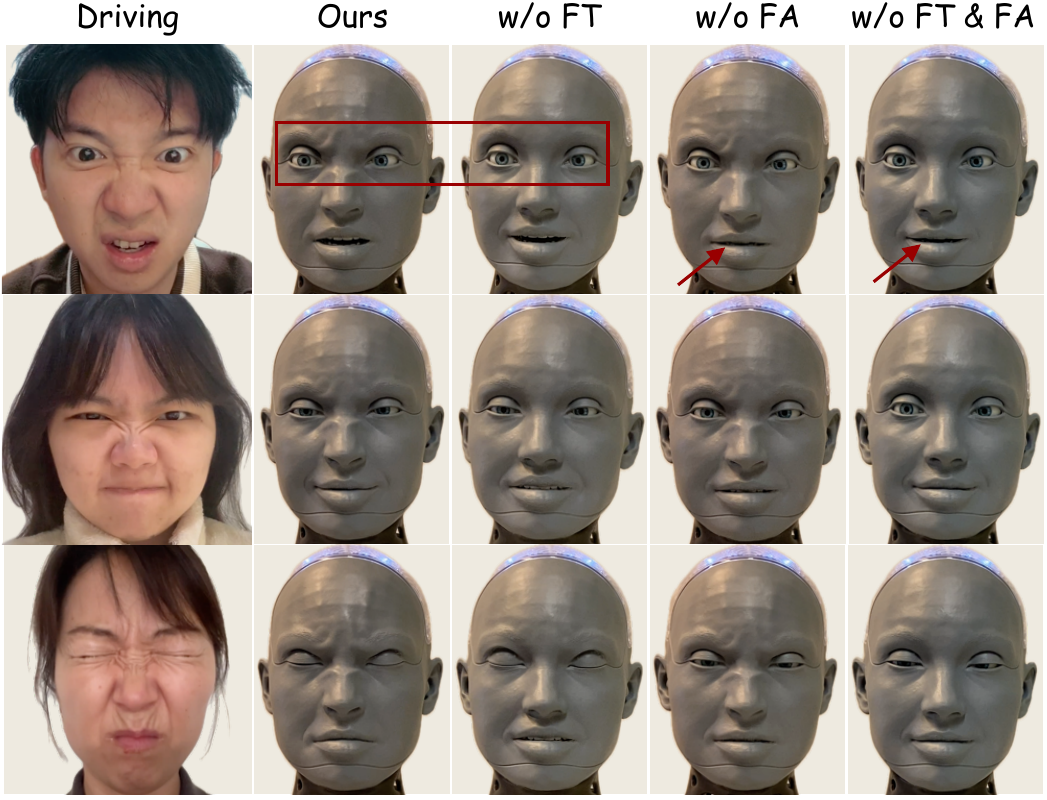}
    \caption{Qualitative ablation studies of fine-tuning and feature-adaptation training.}
    \label{fig:ablations}
\end{figure}

\begin{table}[t]
\centering
\caption{Ablation study in terms of AUR (1–5 scale) and {MAID}.}
\label{tab:ablation}
\begin{tabular}{l c c}
\hline
\textbf{Method} & \textbf{AUR (Mean $\pm$ SD)}~$\uparrow$ &\textbf{MAID}~$\downarrow$\\
\hline
\textbf{X2CNet++} (Ours) & \textbf{4.76} $\pm$ {0.4027}  &\textbf{0.1810}\\
w/o FT            & 4.11 $\pm$ 0.8334 &0.2124\\
w/o FA             & 3.93 $\pm$ 0.7467 &0.2171 \\
w/o FT \& FA             & 3.53 $\pm$ 0.4988 &0.2315 \\
\hline
\end{tabular}
\end{table}

\begin{figure}
    \centering
    \includegraphics[width=1.0\linewidth]{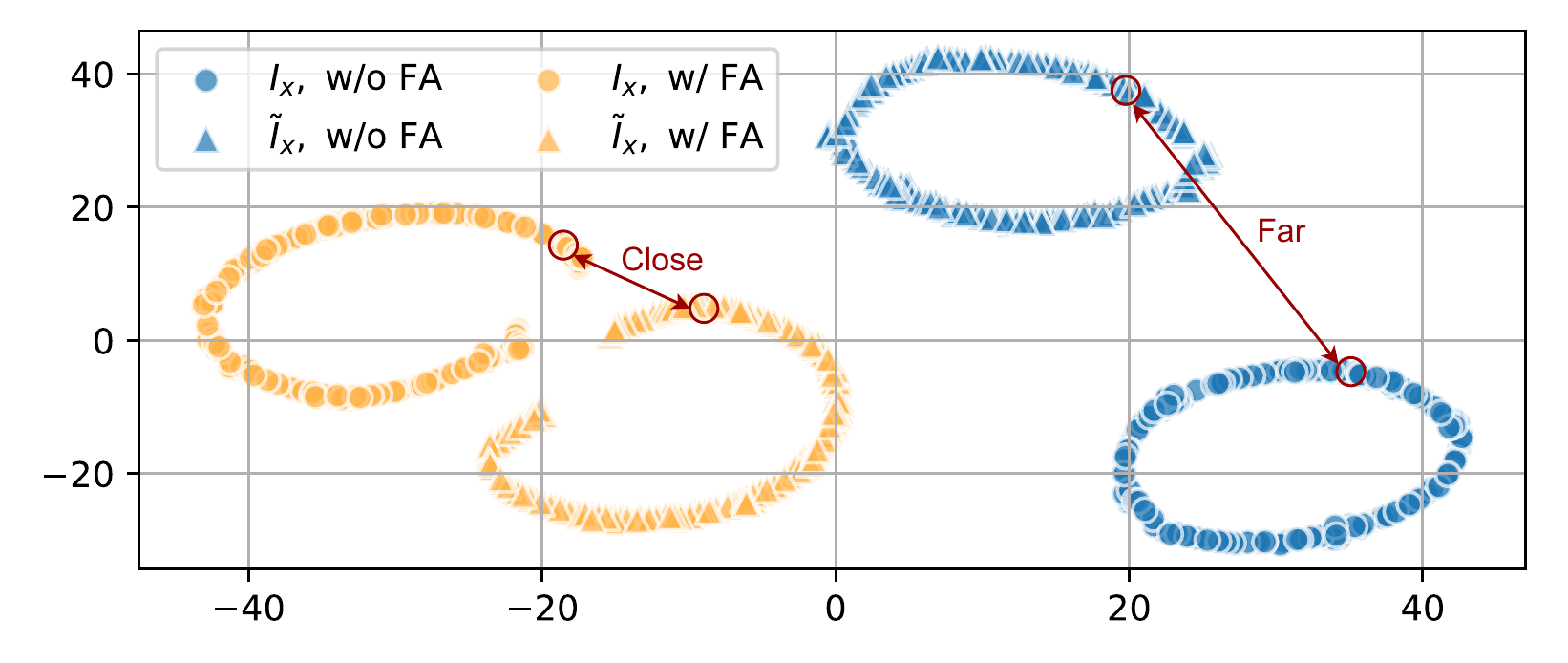}
    \caption{t-SNE visualization of features from selected frames in the X2C dataset and their generated counterparts. The features are extracted from the mapping network’s feature extractor with and without feature-adaptation training (denoted as w/ FA and w/o FA, respectively). }
    \label{fig:featvisualization}
\end{figure}

To validate the effectiveness of fine-tuning the motion transfer module and applying feature-adaptation training to the mapping network, we conduct ablation studies. Qualitative results are shown in Fig.~\ref{fig:ablations}. Without fine-tuning (denoted as w/o FT), the robot fails to reproduce subtle details such as nose wrinkles and frowns; this issue is highlighted in Fig.~\ref{fig:wrinklemissing}, while improvements after fine-tuning appear in Fig.~\ref{fig:wrinkles_solved}.
Without the feature-adaptation training strategy (denoted as w/o FA), the robot struggles to replicate eye and mouth openness, two key cues for recognizing facial expressions. 
To further analyze this, we visualize the features extracted by the mapping network from 300 selected frames (each denoted by $I_x$) in X2C and their generated counterparts $\tilde{I}_x = \mathcal{M}_1(I_x)$ using t-SNE. As illustrated in Fig.~\ref{fig:featvisualization}, with the feature-adaptation strategy (denoted as w/ FA), the generated frames—simulating the outputs of the first stage and serving as inputs to the mapping network during inference—are more closely aligned with the original frames (simulating the training inputs) in feature space, as indicated by red arrows for several example frames. This alignment leads to improved imitation performance, as shown in Fig.~\ref{fig:ablations}.

We also report quantitative results in terms of AUR and MAID on the test set in Table~\ref{tab:ablation}. Performance drops without fine-tuning the motion transfer module or applying feature-adaptation training to the mapping network. Notably, removing both strategies reduces our proposed \textbf{X2CNet++} to the plain X2CNet.

\subsubsection{Real-time Performance}
\begin{table}[h]
\centering
\caption{Latency statistics under different CPU loads. All values are reported in seconds.}
\label{tab:rt_results}
\begin{tabular}{lccccc}
\hline
\textbf{Condition} & \textbf{Mean} & \textbf{Std} & \textbf{P95} & \textbf{P99} & \textbf{Max} \\
\hline
Idle CPU   & 0.0340 &0.0021  & 0.0384 & 0.0447 & 0.0482 \\
50\% Load   & 0.0413  &0.0057 & 0.0482 & 0.0492 & 0.0518 \\
90\% Load     & 0.0459  &0.0061 & 0.0557 & 0.0592 & 0.0695\\
\hline
\end{tabular}
\end{table}

\begin{table}[h]
\centering
\caption{Stage-wise latency budget and observed statistics. All values are reported in seconds.}
\label{tab:rt_budget}
\begin{tabular}{lcccc}
\hline
\textbf{Stage} & \textbf{Budget} & \textbf{Mean} & \textbf{P95} & \textbf{Max} \\
\hline
Model Input Preprocessing & 0.0149 &0.0085 &0.0128 &0.0229 \\
Control Signal Generation  &0.0350 &0.0255 & 0.0266 & 0.0338  \\
Data Transmission & 0.0001 & 3.7e-05 & 5.1e-05 & 0.0004 \\
\hline
Total (E2E)             & 0.0500 & 0.0340 & 0.0384 & 0.0482 \\
\hline
\end{tabular}
\end{table}

We conduct the latency test on a server equipped with an Intel i9-14900K CPU (24 cores, 32 threads) and an NVIDIA RTX 4090 GPU, together with a humanoid robot running the Tritium system\footnote{{https://engineeredarts.com/software/tritium/}}. The results are reported in Table~\ref{tab:rt_results}. As shown, under idle CPU load, the mean latency is 0.0340 seconds and the P99 latency is 0.0447 seconds, indicating that the robot can provide mimicry responses within 0.05 seconds for almost all (99\%) incoming human expressions. Even under 90\% CPU load, the robot still provides mimicry responses within 0.05 seconds on average (mean latency of 0.0459 seconds). These results demonstrate the effectiveness of the proposed real-time humanoid facial expression shadowing system (i.e., \textbf{VividFace}).
We also report the stage-wise latency statistics in Table~\ref{tab:rt_budget}, where the whole processing pipeline is divided into three stages: model input preprocessing (including raw frame cropping and resizing), control signal generation (model inference), and data transmission (sending control values to the robot). As shown, the observed statistics—particularly the mean and P95 latency values—remain within the corresponding budget for each stage.

\section{Conclusion}
This paper presents \textbf{VividFace}, a real-time humanoid facial expression shadowing system that enables humanoid robots to realistically imitate human facial expressions. The key novelty lies in addressing challenges of real-time processing and limited expressiveness caused by the inability to capture subtle expression details. We design a streamlined imitation pipeline that achieves mimicry responses within 0.05 seconds. In addition, we propose fine-tuning the motion transfer module and introducing a feature-adaptation training strategy for the mapping network, resulting in an optimized imitation framework, \textbf{X2CNet++}, which is the core component of \textbf{VividFace}. This allows subtle expression details to be transferred from human to humanoid. Real-world demonstrations, supported by qualitative and quantitative evaluations, validate the effectiveness of \textbf{VividFace} in enhancing humanoid imitation realism and highlight its superiority in delivering natural, smooth, and real-time shadowing for human–robot interaction.

Despite these contributions, \textbf{VividFace} has several limitations. The current design employs a two-stage imitation framework; future work could explore more streamlined architectures to further improve efficiency and performance. Furthermore, the current design assumes a one-on-one human–robot setting, and performance may degrade in multi-person scenarios or in the presence of environmental noise. Future work will investigate the scalability of the system to multi-person settings.











\bibliographystyle{IEEEtran}
\bibliography{mybibfile}

\end{document}